\documentclass[conference]{IEEEtran}
\IEEEoverridecommandlockouts

\usepackage{cite}
\usepackage{amsmath,amssymb,amsfonts}
\usepackage{algorithmic}
\usepackage{graphicx}
\usepackage{textcomp}
\usepackage{bm}
\usepackage{booktabs}
\usepackage[dvipsnames]{xcolor}
\usepackage{color}
\usepackage{hyperref}
\usepackage{url}
\usepackage{subcaption}
\def\BibTeX{{\rm B\kern-.05em{\sc i\kern-.025em b}\kern-.08em
    T\kern-.1667em\lower.7ex\hbox{E}\kern-.125emX}}
\usepackage{pifont,multirow}
\usepackage{booktabs}
\newcommand{\xmark}{\ding{55}}
\newcommand{\eg}{\textit{e.g.}, }
\usepackage{xspace}
\newcommand{\etal}{\textit{et al.}\xspace}
\newcommand{\ie}{\textit{i.e.},\xspace}

\makeatletter
\newcommand{\linebreakand}{%
  \end{@IEEEauthorhalign}
  \hfill\mbox{}\par
  \mbox{}\hfill\begin{@IEEEauthorhalign}
}
\makeatother

\begin{document}

\title{MAVOS-DD: Multilingual Audio-Video Open-Set Deepfake Detection Benchmark}

\author{\IEEEauthorblockN{Florinel Alin Croitoru}
\IEEEauthorblockA{University of Bucharest, Romania}
\and
\IEEEauthorblockN{Vlad Hondru}
\IEEEauthorblockA{
University of Bucharest, Romania}
\and
\IEEEauthorblockN{Marius Popescu}
\IEEEauthorblockA{University of Bucharest, Romania} 
\linebreakand 
\IEEEauthorblockN{Radu Tudor Ionescu}
\IEEEauthorblockA{University of Bucharest, Romania\\
\texttt{raducu.ionescu@gmail.com}}
\and
\IEEEauthorblockN{Fahad Shahbaz Khan}
\IEEEauthorblockA{MBZ University of AI, UAE\\
Link\"{o}ping University, Sweden}
\and
\IEEEauthorblockN{Mubarak Shah}
\IEEEauthorblockA{University of Central Florida, US}
}
\maketitle

\begin{abstract}
We present the first large-scale open-set benchmark for multilingual audio-video deepfake detection. Our dataset comprises nearly 350 hours of real and fake videos across eight languages, with $\mathbf{59\%}$ of data being generated. For each language, the fake videos are generated with several distinct audio and video deepfake generation models, selected based on their quality, via visual inspection. We organize the training, validation and test splits such that only a subset of the chosen generative models and languages are available during training, thus creating several challenging open-set evaluation setups. We perform experiments with various pre-trained and fine-tuned deepfake detectors proposed in recent literature. Our results show that state-of-the-art detectors are not currently able to maintain their performance levels when tested in our open-set scenarios. We publicly release our data and code for non-commercial use at \small{\url{https://github.com/CroitoruAlin/MAVOS-DD}}.
\end{abstract}
\begin{IEEEkeywords}
Deepfake detection, multilingual dataset, multimodal dataset, open-set deepfake detection.
\end{IEEEkeywords}

\begin{figure*}
    \centering
    \includegraphics[width=0.905\textwidth]{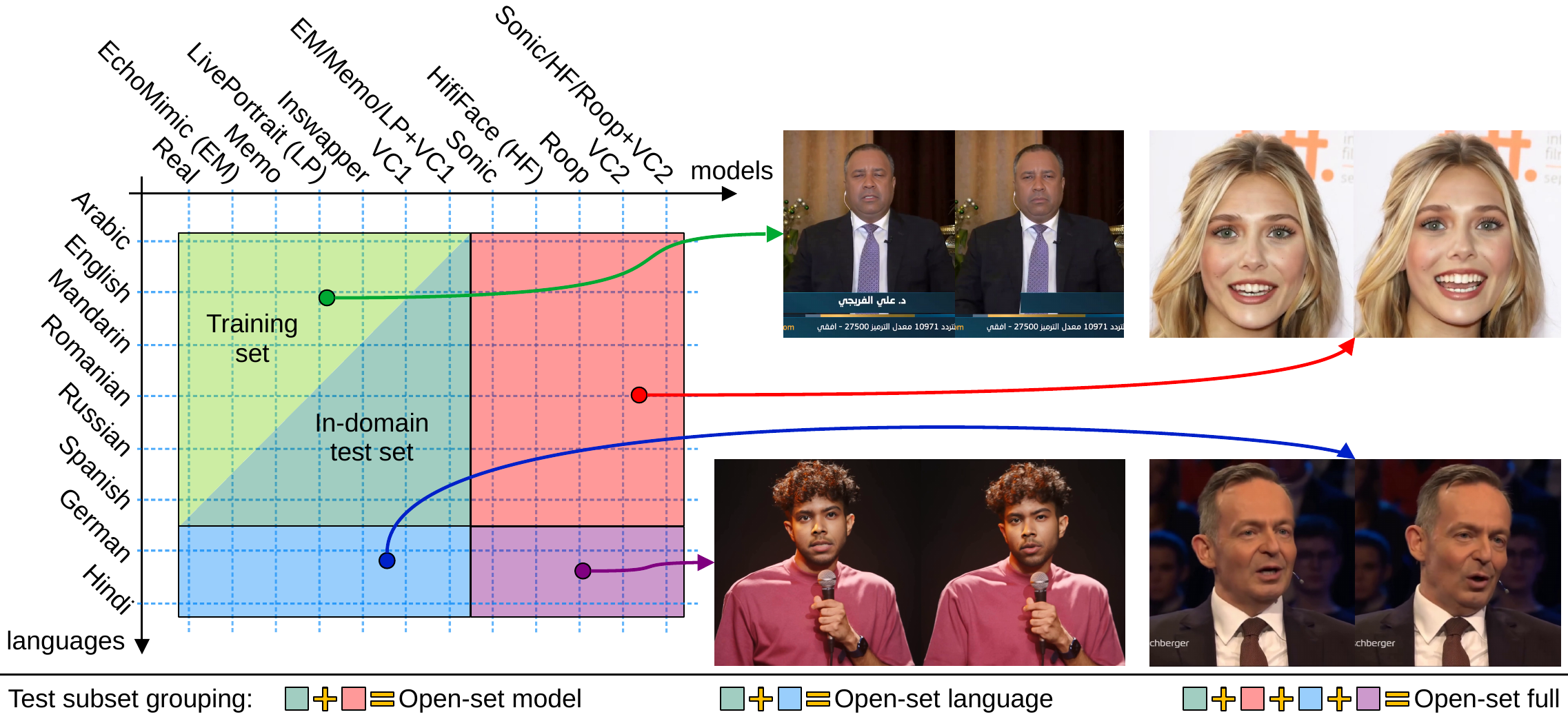}
    \vspace{-0.1cm}
    \caption{In MAVOS-DD, the training set and \emph{in-domain} test set contain real and fake videos sampled from the same distribution, comprising six languages and six generative approaches. The \emph{open-set model} test set extends the in-domain test set with fake samples generated by unseen models (Sonic, HifiFace, Roop, VC2 and Sonic/HifiFace/Roop+VC2). The \emph{open-set language} test set extends the in-domain test set with samples in unseen languages (German and Hindi). The \emph{open-set full} test set adds samples generated by unseen models in unseen languages. One fake sample from each data distribution is shown on the right-hand side. Groups of voice conversion methods (VC1 and VC2) are disjoint across all languages. Best viewed in color.}
    \label{fig_teaser}
    \vspace{-0.4cm}
\end{figure*}

\vspace{-0.1cm}
\section{Introduction}
\vspace{-0.1cm}

The rapid progress in image, audio and video synthesis technologies has enabled the creation of realistic visual content from textual descriptions \cite{Croitoru-TPAMI-2023,ramesh-ArXiv-2022,Saharia-NeurIPS-2022,Rombach-CVPR-2022,Nichol-ICML-2022}, as well as the convincing manipulation of people's identities \cite{li-CVPR-2020,chen-ACMICM-2020,joo-IC3DV-2021,nirkin-ICCV-2019} and expressions \cite{wang-ICAFGR-2024,zheng-CVPR-2022,hong-CVPR-2022,Chen-AAAI-2025,Xu-ArXiv-2024a,Xu-NeurIPS-2024,Tian-ArXiv-2024}. This has led to a surge of innovative applications across various industries, including marketing and film making. However, these breakthroughs have also fueled the rise of malicious uses, particularly in generating deceptive synthetic audio-visual content, commonly known as deepfakes \cite{Croitoru-Arxiv-2024}. Alarmingly, a recent report shows that the incidence of deepfake-related fraud increased by a factor of 10 between 2022 and 2023\footnote{\href{https://sumsub.com/blog/sumsub-experts-top-kyc-trends-2024/}{Sumsub Expert Roundtable: The Top KYC Trends Coming in 2024}}. In this landscape, the ability to reliably identify forged video material is more crucial than ever.


A significant body of research has emerged in response to the rising number of deepfake-related manipulation and fraud cases, aiming to detect manipulated content using advanced deep learning techniques, such as convolutional neural networks \cite{raza-2023-CVPR,cozzolino-CVPR-2023,kihal-MTA-2023,Ciamarra-WACVW-2024,Lanzino-CVPRW-2024,Ba-AAAI-2024}, transformers \cite{zhou-ICCV-2021,oorloff-CVPR-2024, salvi-JI-2023, Ilyas-ASC-2023, Zhang-ACMTMCCA-2024,nie-ACMMM-2024}, and hybrid approaches \cite{bonettini-ICPR-2021,wang-AAAI-2023,coccomini-ICIAP-2022,Guan-NeurIPS-2022,zheng-ICCV-2021,choi-CVPR-2024}. These methods have achieved remarkable results, often surpassing 99\% accuracy on existing benchmarks \cite{Croitoru-Arxiv-2024}, such as Celeb-DF \cite{yuezun-CVPR-2020} and FaceForensics++ \cite{andreas-ICCV-2019}. Nevertheless, most evaluations are carried out in controlled environments where the synthetic and authentic samples in training and testing originate from the same video manipulation tools. This in-domain evaluation setup significantly inflates detection performance and fails to represent real-world conditions, where neither the manipulated technique nor the subject is known in advance.

To address this gap, we propose a new benchmark for evaluating audio-video deepfake detection models in a multilingual open-world setup. Our benchmark, MAVOS-DD, comprises over 51K fake and 35K real videos, totaling nearly 350 hours of video across eight languages: Arabic, English, German, Hindi, Mandarin, Romanian, Russian and Spanish. The fake samples are generated by several state-of-the-art deepfake generation methods based on different approaches: talking head (EchoMimic \cite{Chen-AAAI-2025}, Memo \cite{Zheng-ArXiv-2024}, Sonic \cite{Ji-CVPR-2025}), portrait animation (LivePortrait \cite{Guo-ArXiv-2024}), face swap (Inswapper\footnote{\url{https://github.com/deepinsight/insightface}}, HifiFace \cite{Wang-IJCAI-2021}, Roop\footnote{\url{https://github.com/s0md3v/roop}}), and voice conversion (FreeVC \cite{li2022freevchighqualitytextfreeoneshot}, KNN-VC \cite{baas2023knnvc}, OpenVoice \cite{qin2024openvoiceversatileinstantvoice}, XTTSv2 \cite{casanova2024xttsmassivelymultilingualzeroshot}, YourTTS \cite{Casanova-ICML-2022}). As shown in Figure \ref{fig_teaser}, we create a multi-perspective open-set benchmark that comprises video, audio and audio-video manipulations. For audio manipulation, multiple voice conversion (VC) methods are required to cover all languages. We group VC methods into two groups, VC1 and VC2, such that methods in each group are disjoint for every language. The training set comprises audio-video samples in six languages (excluding German and Hindi), where the fake samples are generated via eight approaches (EchoMimic, Memo, LivePortrait, Inswapper, VC1, EchoMimic+VC1, Memo+VC1, and LivePortrait+VC1).
We establish an in-domain (closed) test set that is sampled from the same distribution as the training data. In addition, we create three open-set test sets: (i) \emph{open-set model} extends
the in-domain test set with fake samples generated by unseen models; (ii) \emph{open-set language} adds German and Hindi samples to the in-domain test data; (iii) \emph{open-set full} adds samples generated by unseen models in German and Hindi. 

We perform extensive experiments using both pre-trained and fine-tuned deep fake detectors \cite{oorloff-CVPR-2024,Smeu-CVPR-2025,Zou-ICASSP-2024,xu-ICCV-2023}, analyzing their performance on both in-domain and open-set scenarios. While these models work well under in-domain conditions, one of them surpassing an accuracy threshold of $90\%$, their effectiveness drops significantly in the open-set setups. The reported performance gaps highlight a critical limitation of current deepfake detection models, namely the poor generalization across deepfake generation models and languages. 

In summary, our contribution is twofold:
\begin{itemize}
\item 
We present MAVOS-DD, a comprehensive multilingual open-set benchmark for audio-video deepfake detection, encompassing almost 350 hours of authentic and synthetic videos across eight languages.
\item We conduct a thorough evaluation of state-of-the-art deepfake detectors, uncovering substantial performance degradation when models are tested in open-world setups, thereby emphasizing the need for more robust and generalizable detection techniques.
\end{itemize}

\vspace{-0.1cm}
\section{Related Work}
\vspace{-0.1cm}
The field of deepfake generation has seen significant advancements in recent years \cite{Croitoru-Arxiv-2024}, particularly with the rise of diffusion models \cite{Croitoru-TPAMI-2023,Ho-NeurIPS-2020,Rombach-CVPR-2022,Saharia-NeurIPS-2022,song-NeurIPS-2019}. In parallel, considerable research has been devoted to developing effective detection techniques \cite{Croitoru-Arxiv-2024,oorloff-CVPR-2024,Zou-ICASSP-2024,xu-ICCV-2023} to counter the negative effects of deepfake media. In addition, substantial efforts have been made to construct datasets for deepfake detection \cite{andreas-ICCV-2019,dolhansky-arXiv-2020,jiang-CVPR-2020,yuezun-CVPR-2020,khalid-NeurIPS-2021}, thereby facilitating research in this domain.

\noindent
\textbf{Audio-visual deepfake detection.} Traditional deepfake detection methods are unimodal, focusing solely on either visual artifacts, \eg abnormal facial textures \cite{Lanzino-CVPRW-2024,kingra-FSIDI-2022,fang-ACMTPS-2025} and inconsistent lighting \cite{gerstner-CVPR-2022}, or audio inconsistencies, \eg speech prosody \cite{logan-USENIX-2022, wang-Interspeech-2023, attorresi-ICPR-2023}, frequency patterns \cite{sriskandaraja-interspeech-2016,yang-TIFS-2020,fan-SC-2023,xue-IWDDAM-2022}, and voice cloning artifacts \cite{martin-IJCAI-2023,gao-INTERSPEECH-2021}. With generation methods becoming more capable, it is essential to leverage both visual and auditory modalities to improve the robustness and reliability of the forgery detection models \cite{oorloff-CVPR-2024,Zou-ICASSP-2024,xu-ICCV-2023}. Aside from unimodal cues, utilizing multimodal (audio-visual) information can naturally capitalize on the misalignment between the two modalities by examining if the audio and video signals are coherent and temporally aligned, \eg in terms of lip movements \cite{agarwal-CVPR-2020,zhou-ICCV-2021} or facial expressions \cite{haliassos-CVPR-2022}.

\begin{table*}[t]
  \centering
      \caption{Comparison between MAVOS-DD and other video and audio-video (multimodal) datasets. MAVOS-DD is the largest dataset for multilingual audio-video open-set deepfake detection.}
      \vspace{-0.15cm}
  \fontsize{8}{9}\selectfont{
  \begin{tabular}{lrrrrrcccc}
    \toprule
    \multirow{2.5}{*}{Dataset}     & \multicolumn{2}{c}{\multirow{1}{*}{File count}} & \multicolumn{3}{c}{\multirow{1}{*}{Length (h)}} & \multirow{2.5}{*}{\#Generative methods} & \multirow{2.5}{*}{\rotatebox[origin=c]{0}{\#Languages}} & 
\multirow{2.5}{*}{\rotatebox[origin=c]{0}{Open-set}} & \multirow{2.5}{*}{\rotatebox[origin=c]{0}{Multimodal}} \\
\cmidrule(lr){2-3}
\cmidrule(lr){4-6}
        & \multirow{1.0}{*}{\#Real} & \multirow{1.0}{*}{\#Fake} & \multirow{1.0}{*}{Real} & \multirow{1.0}{*}{Fake} & \multirow{1.0}{*}{Total} &  & & & \\
    \midrule
    FaceForensics++ \cite{andreas-ICCV-2019} & 1,000 & 4,000 & 4.7  & 17.0 & 21.7 & 4 & 0 & \textcolor{red}{\xmark} & \textcolor{red}{\xmark} \\
    DFDC \cite{dolhansky-arXiv-2020} & 23,654 & 104,500 & 64.4 & 288.9 & 353.3 & 5 & 0 & \textcolor{red}{\xmark} & \textcolor{red}{\xmark} \\
    DeeperForensics \cite{jiang-CVPR-2020} & 50,000 & 10,000 & 46.3 & 116.7 & 163.0 & 1 & 0 & \textcolor{red}{\xmark} & \textcolor{red}{\xmark} \\
    ForgeryNet \cite{he-CVPR-2021} & 99,630 & 121,617 & 13.3 & 13.5 & 26.8 & 15 & 0 & \textcolor{red}{\xmark} & \textcolor{red}{\xmark} \\
    Celeb-DF \cite{yuezun-CVPR-2020} & 590 & 5,639 & 2.1 & 20.4 & 22.5 & 1 & 0 & \textcolor{red}{\xmark} & \textcolor{red}{\xmark} \\
    WildDeepfake \cite{zi-ACM-2020} & 3,805 & 3,509 & - & - & 10.9 & N/A & 0 & \textcolor{red}{\xmark}  & \textcolor{red}{\xmark} \\

    FakeAVCeleb \cite{khalid-NeurIPS-2021} & 500 & 19,500 & 1.1 & 41.2 & 42.3 & 3 & 1 & \textcolor{red}{\xmark} & \textcolor{ForestGreen}{\checkmark} \\
    
    DeepSpeak \cite{barrington-arXiv-2024} &  6,226 & 6,799 & 17.0 & 26.0 & 44.0 & 10 & 1 & \textcolor{red}{\xmark} & \textcolor{ForestGreen}{\checkmark} \\

    BioDeepAV \cite{Croitoru-Arxiv-2024} &  2,010 & 1,693 & 3.2 & 8.2 & 11.4 & 4 & 1 & \textcolor{red}{\xmark} & \textcolor{ForestGreen}{\checkmark} \\

    Deepfake-Eval-2024 \cite{chandra-arXiv-2025} & 1,072 & 964 & 28.9 &  16.2 & 45.1 & N/A & 49 & \textcolor{red}{\xmark} & \textcolor{ForestGreen}{\checkmark} \\
    PolyGlotFake \cite{Hou-ICPR-2024} & 766 & 14,472 & 2.6 &  48.3 & 50.9 & 10 & 7 & \textcolor{red}{\xmark} & \textcolor{ForestGreen}{\checkmark} \\
    \midrule
    MAVOS-DD (ours) & 35,557  & 51,481 & 127.9 & 221.9 & 349.8 & 15  & 8 & \textcolor{ForestGreen}{\checkmark} & \textcolor{ForestGreen}{\checkmark}  \\
    \bottomrule
  \end{tabular}
  }

    \label{table_datasets_comparison}
  \vspace{-0.3cm}
\end{table*}

Early works on audio-visual deepfake detection used convolutional architectures \cite{raza-2023-CVPR,cozzolino-CVPR-2023,kihal-MTA-2023}. For example, Multimodaltrace \cite{raza-2023-CVPR} extracts separate features from audio and video with residual blocks, fuses the resulting representations and further processes them to make the final prediction. Kihal and Hamza~\cite{kihal-MTA-2023} also employ individual CNN-based feature extractors, but use a Random Forest model to predict the final label.
Recent works opted for architectures that leverage transformers, not only because of their higher performance, but also because of the inherent mechanism that enables fusing the information from two modalities using cross-attention modules \cite{zhou-ICCV-2021,oorloff-CVPR-2024, salvi-JI-2023, Ilyas-ASC-2023, Zhang-ACMTMCCA-2024,nie-ACMMM-2024}. Zhou and Lim~\cite{zhou-ICCV-2021} detect inconsistencies between the two modalities (focusing on lip movements and speech) by aligning their low-level latent representations and fusing them through a cross-modal attention mechanism. Nie~\etal~\cite{nie-ACMMM-2024} employ two pre-trained frozen ViTs~\cite{Dosovitskiy-ICLR-2021} to extract features, with only the \texttt{[CLS]} tokens being used for classification. To bridge the gap between modalities, the audio information is integrated into the visual tokens using an audio-distilled cross-modal interaction module. Furthermore, the authors try to detect high-frequency forgery artifacts by biasing the queries, keys and values with learnable parameters.
  
\noindent
\textbf{Audio-visual deepfake datasets.} While the advancement of deepfake generation methods has led to the development of detection methods to defend against deepfakes, it has also driven the need for extensive datasets. In the beginning, datasets comprising data from a single modality were created for both visual (image and video) data \cite{dang-CVPR-2020, dolhansky-arXiv-2020, he-CVPR-2021, chen-arXiv-2024, andreas-ICCV-2019, li-CVPR-2020, yuezun-CVPR-2020,zi-ACM-2020} and audio data \cite{wang-CSL-2020, Liu-TASLP-2021}. Nevertheless, with the rise of multimodal models, the availability of audio-visual datasets \cite{korshunov-arXiv-2018,khalid-NeurIPS-2021,chandra-arXiv-2025,barrington-arXiv-2024} has become essential. 

We present a comprehensive comparison of MAVOS-DD with other video and multimodal datasets in Table~\ref{table_datasets_comparison}. DFDC \cite{dolhansky-arXiv-2020} is among the largest video dataset for deepfake detection. However, multimodal datasets, such as FakeAVCeleb \cite{khalid-NeurIPS-2021} and Deepfake-Eval-2024 \cite{chandra-arXiv-2025} are not as large. FakeAVCeleb \cite{khalid-NeurIPS-2021} is based on two face swapping methods and a facial reenactment method for their synthetic English-speaking videos. 
While DeepSpeak \cite{barrington-arXiv-2024} tries to excel by employing 10 generative methods, Deepfake-Eval-2024 \cite{chandra-arXiv-2025} stands out by having videos in 49 languages, although $80\%$ of its content is in English.

One of the main limitations of the deepfake detection methods is their ability to generalize to synthetic samples generated with different methods. To this end, MAVOS-DD contains samples obtained with a variety of generative methods to facilitate training robust detection models, but also to thoroughly evaluate their ability to generalize to unseen methods. Moreover, with only two exceptions \cite{chandra-arXiv-2025,Hou-ICPR-2024} from concurrent literature, existing datasets do not focus on the multilingual aspect of audio-visual content. Chandra~\etal~\cite{chandra-arXiv-2025} collect a dataset from the web, so there is no control over the generative methods and languages. In contrast, our dataset enables an open-set evaluation in terms of both generative models and languages. Furthermore, our dataset comprises 14$\times$ more deepfake content (222 hours vs.~16 hours), which enables the training of very deep models with higher generalization capacity. Although their videos span 49 languages, $80\%$ of all videos are in English (each other language representing less than $0.5\%$ of the dataset). In this regard, MAVOS-DD provides a more even distribution across languages (see Figure~\ref{fig:sub-langs}). PolyGlotFake~\cite{Hou-ICPR-2024} contains 766 real and 14,472 fake videos, resulting in an imbalance ratio of approximately 1:18.5. In contrast, MAVOS-DD yields a more balanced ratio of roughly 1:1.4. Moreover, PolyGlotFake employs only two video manipulation methods, with the most recent being VideoRetalking~\cite{Cheng-SIGGRAPH-2022}. In MAVOS-DD, we incorporate more recent and diverse manipulation techniques, including Sonic~\cite{Ji-CVPR-2025} and LivePortrait~\cite{Guo-ArXiv-2024}, offering a more up-to-date and challenging benchmark. Overall, the comparison in Table~\ref{table_datasets_comparison} shows that MAVOS-DD is the largest dataset for multilingual audio-video open-set deepfake detection. 

\vspace{-0.1cm}
\section{Dataset}
\label{sec:dataset}
\vspace{-0.1cm}

\begin{figure*}[!t]
\begin{subfigure}[t]{.48\textwidth}
  \centering
  \includegraphics[width=.98\linewidth]{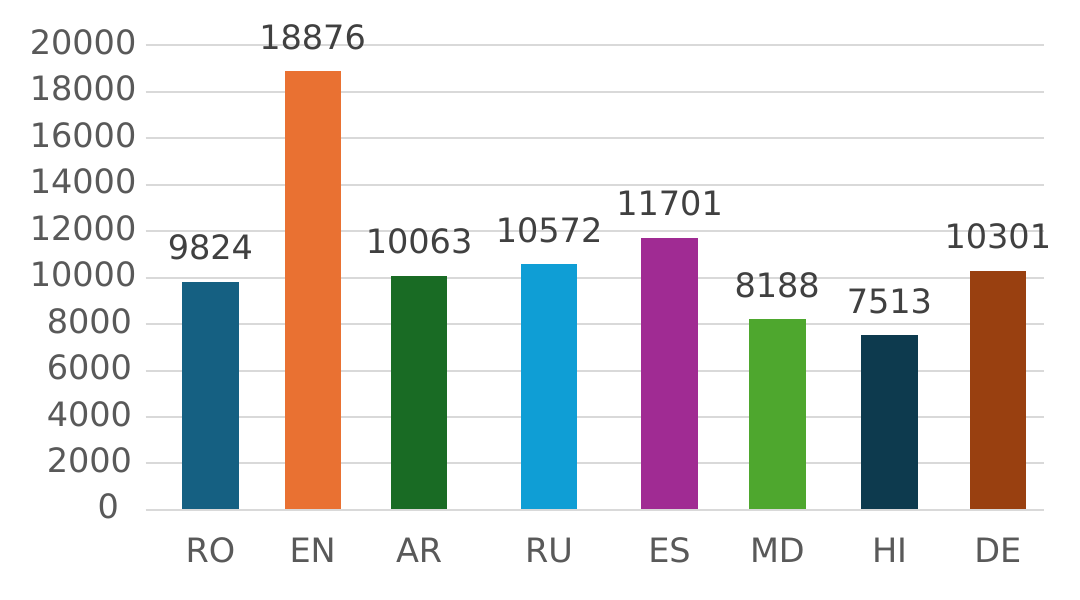}  
  \vspace{-0.1cm}
  \caption{Number of real and deepfake videos per language.}
  \label{fig:sub-langs}
\end{subfigure}
\hfill
\begin{subfigure}[t]{.48\textwidth}
  \centering
\includegraphics[width=.98\textwidth]{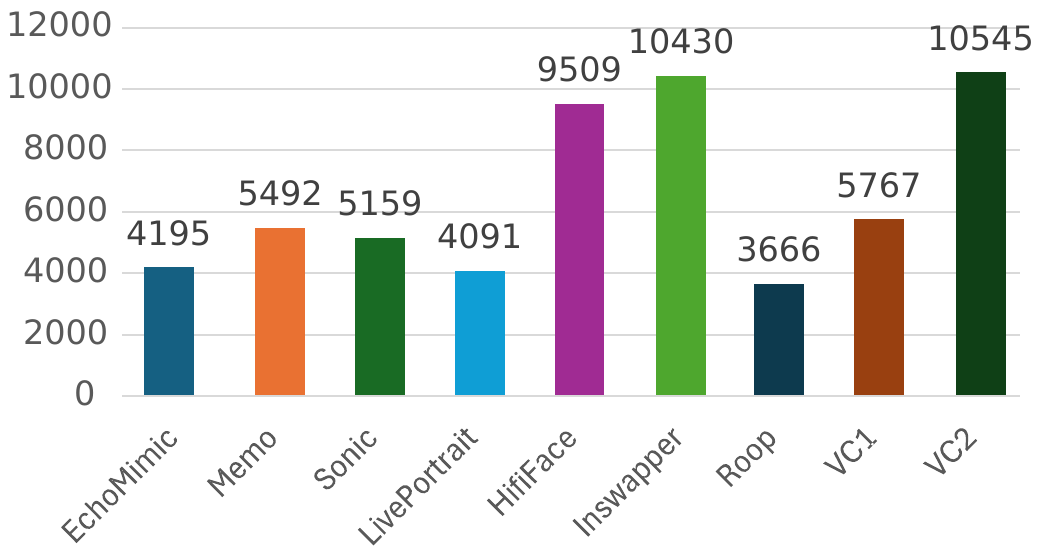}  
 \vspace{-0.1cm}
  \caption{Number of deepfake videos generated with each method.}
  \label{fig:sub-models}
\end{subfigure}
\vspace{-0.05cm}
\caption{Distribution of videos per language and per generative method. MAVOS-DD comprises videos in eight languages, generated with various methods. The languages are coded as follows: Arabic (AR), English (EN), German (DE), Hindi (HI), Mandarin (MD), Romanian (RO), Russian (RU) and Spanish (ES).}
\label{fig_ablation}
  \vspace{-0.3cm}
\end{figure*}

\noindent
\textbf{Overview.}
Our main contribution is MAVOS-DD, a large-scale deepfake dataset consisting of 87,038 real and synthetic videos, totaling 350 hours of content across eight different languages. The synthetic content is generated using a wide variety of video methods (EchoMimic~\cite{Chen-AAAI-2025}, Memo~\cite{Zheng-ArXiv-2024}, Sonic~\cite{Ji-CVPR-2025}, LivePortrait~\cite{Guo-ArXiv-2024}, Inswapper, HifiFace~\cite{Wang-IJCAI-2021}, and Roop) and audio methods (FreeVC \cite{li2022freevchighqualitytextfreeoneshot}, KNN-VC \cite{baas2023knnvc}, OpenVoice \cite{qin2024openvoiceversatileinstantvoice}, XTTSv2 \cite{casanova2024xttsmassivelymultilingualzeroshot}, YourTTS \cite{Casanova-ICML-2022}). The VC methods are divided into two groups, denoted as VC1 (FreeVC, XTTSv2) and VC2 (OpenVoice, YourTTS, KNN-VC). The deepfake methods cover a broad range of key generative tasks: talking-head generation (EchoMimic, Memo, Sonic), facial expression transfer (LivePortrait), face swapping (Inswapper, HifiFace, Roop), voice conversion (FreeVC, KNN-VC, OpenVoice, XTTSv2, YourTTS), joint talking-head and voice conversion (EchoMimic/Memo+VC1, Sonic+VC2), joint facial expression transfer and voice conversion (LivePortrait+VC1), and joint face swapping and voice conversion (HifiFace/Roop+VC2). This coverage ensures a diverse and realistic set of generated videos, comprising three kinds of deepfakes: video, audio, and multimodal (audio-video). The main reason for using recent generative methods is to create a challenging dataset. Yet, another level of complexity is added through the fact that the audio-video samples cover eight languages: Arabic (AR), English (EN), German (DE), Hindi (HI), Mandarin (MD), Romanian (RO), Russian (RU) and Spanish (ES). We present the video distribution per language and per generative method in Figure~\ref{fig:sub-langs} and Figure~\ref{fig:sub-models}, respectively. Note that real videos are naturally included in the distribution of videos per language, but not in the distribution of videos per generative method. The distribution per language is influenced by the number of real videos collected for each language, while the distribution per method is influenced by the speed of each generative method. The total time required to generate all videos included in MAVOS-DD amounts to roughly 98 days (time measured on a computer with an Intel i9-14900K CPU with 192 GB of RAM and an Nvidia RTX 4090 GPU with 24 GB of VRAM).

We define official training, validation, and test splits for various evaluation scenarios, as illustrated in Figure~\ref{fig_teaser}. The first scenario, referred to as \emph{in-domain} evaluation, uses a test set comprising the same languages and generative methods as the training set. The second and third scenarios, namely \emph{open-set model} and \emph{open-set language}, expand the in-domain test set to include samples generated by unseen models or unseen languages, respectively. The final scenario, called \emph{open-set full}, includes samples generated by unseen models in unseen languages, presenting the most challenging evaluation setting. We present detailed statistics about MAVOS-DD and its splits in Table~\ref{table_statistics}. The training and validation splits do not include videos in German or Hindi, as these languages are reserved exclusively for the test set to support {open-set} evaluation. Overall, the number of real and fake samples is relatively balanced. However, the \emph{open-set model} and \emph{open-set full} splits contain a slightly larger number of fake samples, as they comprise synthesized videos from three additional generative methods that are not present in the training set, as illustrated in Figure~\ref{fig_teaser}.

\begin{table*}[t]
  \centering
      \caption{Number of real and fake videos included in the training, validation and test splits of MAVOS-DD. The test data is divided into four subsets, which generate an in-domain evaluation scenario and three open-set evaluation scenarios. The core set includes six languages (Arabic, English, Mandarin, Romanian, Russian, Spanish) and four methods (EchoMimic, Memo, LivePortrait, Inswapper). The extra languages are German and Hindi. The extra models are Sonic, HifiFace, Roop and VC2. The length (in hours) of the real and fake content in each split is reported in the last column.}
        \vspace{-0.15cm}
  \fontsize{8}{9}\selectfont{
  \begin{tabular}{llccccccc}
    \toprule
    \multicolumn{2}{l}{\multirow{2.5}{*}{Split}} & \multirow{2.5}{*}{Video type} & \multicolumn{4}{c}{File count} &\multirow{2.5}{*}{Total count} & \multirow{2.5}{*}{Total length (h)}\\
    \cmidrule(lr){4-7}
    & &  & Core set & Extra langs & Extra models & Extra models \& langs &  & \\
    \midrule
    \multicolumn{2}{l}{\multirow{2}{*}{Train}} & Real & 10,297 & 0 &  0 & 0  & 10,297& 38.5\\
    & & Fake & 11,073 & 0 & 0 & 0 & 11,073 & 49.1\\
    \midrule
    \multicolumn{2}{l}{\multirow{2}{*}{Validation}} & Real & 1,715 & 0 & 0 & 0 & 1,715& 6.4\\
    & & Fake & 2,180 & 0 & 0 & 0 & 2,180 & 9.5\\
    \midrule
    \multirow{9.5}{*}{Test} & \multirow{2}{*}{In-domain} & Real & 5,185 & 0 &  0 & 0 & 5,185 & 19.3\\
    & & Fake & 7,223 & 0 & 0 & 0 & 7,223 & 32.7\\
    \cmidrule{2-9}
    & \multirow{2}{*}{Open-set language} & Real & 5,185 & 7,998 &  0& 0 & 13,183& 46.4\\
    & & Fake & 7,223 & 5,332 & 0 & 0 & 12,555 & 60.9 \\
    \cmidrule{2-9} 
    & \multirow{2}{*}{Open-set model} & Real & 5,185 & 0 &  10,362 & 0 & 15,547& 56.1\\
    & & Fake & 7,223 & 0 & 21,189 & 0 & 28,412 &  100.6\\
    \cmidrule{2-9}
    & \multirow{2}{*}{Open-set full} & Real & 5,185 & 7,998 &  10,362 & 0 & 23,545& 83.1\\
    & & Fake &  7,223 & 5,332 & 21,189 & 4,484 & 38,228 & 163.3\\
    \bottomrule
  \end{tabular}
  }

      \label{table_statistics}
  \vspace{-0.1cm}
\end{table*}

\noindent
\textbf{Real videos.} We collect real videos from YouTube, primarily sourcing content from public news channels or street interviews in each target language (such as EasyLanguages\footnote{\url{https://www.easy-languages.org/}}). Additionally, we include videos from well-known channels specific to each country and language. 
After obtaining the videos, we apply the TalkNet active speaker detection model~\cite{tao-ACM-2021} to segment the videos into shorter clips, each featuring a single speaking individual. As the process to acquire the videos and split them into smaller videos is automatic, there are some instances where the videos do not contain any humans, \ie~faces. In order to filter these out, for each video, we apply a face detector \cite{Jocher-github-Ultralytic} on individual frames (using a step of 15 frames) and eliminate those videos that do not have a face for more than half of the evaluated frames. The final dataset comprises 35,557 high-quality videos, with resolutions ranging from $256\times256$ to $1920\times1080$, amounting to a total of 128 hours of real content.
       
         
\noindent
\textbf{Deepfake videos.} Deepfake generation typically involves a source identity image, representing the face that is manipulated by the generative model. We take these identities from multiple sources in our experiments. The first source is a set of 500 portraits generated by us using FLUX\footnote{\url{https://github.com/black-forest-labs/flux}}. We use the simple text prompt ``A portrait of a man/woman'', as it consistently produces high-quality images without compromising output diversity. For the diffusion process, we set the number of denoising steps to $50$ and use a guidance scale of $3.5$. Additionally, we supplement the generated portraits with real identities from well-established face datasets, specifically FFHQ~\cite{Karras-TPAMI-2021} and CelebAMask-HQ~\cite{Lee-CVPR-2020}, along with identities found in our real videos. These datasets have disproportional dimensions, but we sample subsets from each to ensure an almost uniform distribution across datasets.

\begin{table}[t]
    \setlength\tabcolsep{0.5em}
    \centering
    \caption{MAVOS-DD splits across real and fake modalities. Legend: RV -- real video; FV -- fake video; RA -- real audio; FA -- fake audio.}
         \vspace{-0.15cm}
    \fontsize{8}{9}\selectfont{
    \begin{tabular}{lcccc}
    \toprule
     & RVRA & FVRA & RVFA & FVFA\\
    \midrule
    Sample count & 35,557 & 35,169 & 8,939 & 7,373 \\
    Length (h) & 128 & 160 & 32 & 29 \\
    \bottomrule
    \end{tabular}
    }
        
            \label{av-fake-distribution}
    \vspace{-0.3cm}
\end{table}
The talking-head generation is performed with EchoMimic, Memo and Sonic. We provide these models with a portrait image, sampled from the previously described set, and an audio signal containing a person speaking. The audio also originates from the real video set described earlier. The result is a video in which the person from the portrait image utters the speech from the audio file. We emphasize that the models not only manage lip synchronization, but also effectively generate head movements and facial expressions required for this task. Furthermore, we observe that Memo and Sonic perform consistently well across multiple languages, while EchoMimic struggles with languages other than English and Mandarin. For this reason, we individually fine-tune EchoMimic on additional languages, such as Romanian and Arabic, before using it for generation. We use 1,000 real videos for each language and train the model for 10 epochs. Finally, we synthesize over 14,000 videos using talking-head generation methods, resulting in more than 83 hours of fake content. All videos are generated at a consistent resolution of $512\times512$ pixels.

For facial expression manipulation, we employ LivePortrait~\cite{Guo-ArXiv-2024}. This model can transfer facial movements (eyes, lips, and expressions) from a driving video to a source image or video. However, we observe a noticeable drop in quality when the person in the driving video is not directly facing the camera. Additionally, while lip synchronization is handled effectively, the transfer of eye movements and facial expressions is less effective. To address these limitations, we restrict our use to front-facing driving videos and focus only on lip synchronization. As a result, only the movements of the lips are synthesized in the generated samples, while all other facial attributes in the source video remain unchanged. The audio of the resulting video is taken from the driving video, to ensure alignment between the lips and the information spoken in the audio. We select front-facing driving videos from the set generated using talking-head synthesis, as these are primarily created from portrait images, and verified for the front-facing property. The source videos are represented by the real videos collected from YouTube. We generate over 4,000 videos using this method, resulting in more than 20 hours of fake content. The generated videos inherit the resolution of the source (real) videos, as the only changed aspect is the movement of the lips.

\begin{figure}[t]
    \centering
    \includegraphics[width=1.0\linewidth]{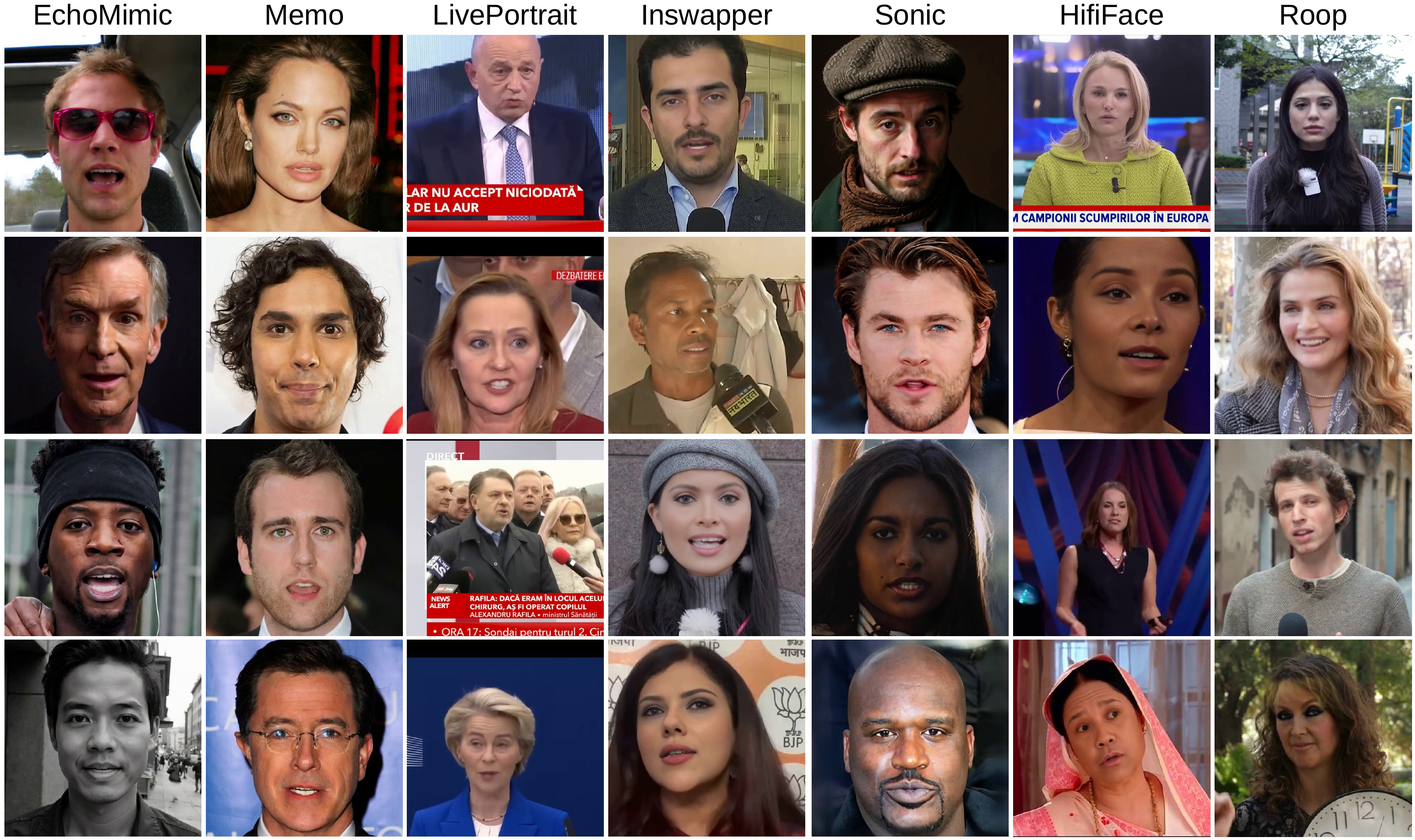}
    \vspace{-0.5cm}
    \caption{Fake video frames generated by several deepfake generation methods. Best viewed in color.}
    \label{fig_qualitative}
    \vspace{-0.3cm}
\end{figure}

The face swapping is performed with Inswapper, HifiFace and Roop. Face swapping works by pasting the identity from a source image to a target video, while keeping the attributes that are not specific to the identity (facial expression, lip movement) unchanged. For the source images, we use portraits from the previously described dataset, which includes both synthetic and real identities. The target videos are selected from the collected set of real YouTube videos. Following face swapping, we apply GFPGAN~\cite{Wang-CVPR-2021} for face restoration to enhance visual quality. We generate over 22,000 videos using this deepfake method, totaling 85 hours of fake content. The resolution of the resulting videos matches that of the target (real) videos.

\begin{table*}
  \centering
  \caption{Results obtained by pre-trained and fine-tuned versions of AVFF~\cite{oorloff-CVPR-2024}, MRDF~\cite{Zou-ICASSP-2024}, TALL~\cite{xu-ICCV-2023} and AVH-Align~\cite{Smeu-CVPR-2025} on the MAVOS-DD official test sets: in-domain, open-set model, open-set language and open-set full. The best and second-best results on each column are highlighted in {\color{blue}\textbf{bold blue}} and {\color{RedOrange}orange}, respectively. 
  }
  \vspace{-0.15cm}
  \fontsize{8}{9}\selectfont{
  \begin{tabular}{lccccccccccccc}
    \toprule
      \multirow{2.5}{*}{Method} & \multirow{2.5}{*}{{Fine-tuned}} & \multicolumn{3}{c}{\multirow{1}{*}{In-domain}} & \multicolumn{3}{c}{\multirow{1}{*}{Open-set model}} & \multicolumn{3}{c}{\multirow{1}{*}{Open-set language}} & \multicolumn{3}{c}{\multirow{1}{*}{Open-set full}} \\
     \cmidrule(lr){3-5}
     \cmidrule(lr){6-8}
     \cmidrule(lr){9-11}
     \cmidrule(lr){12-14}
    &  & \multirow{1}{*}{mAP} & \multirow{1}{*}{AUC} & \multirow{1}{*}{acc} & \multirow{1}{*}{mAP} & \multirow{1}{*}{AUC} & \multirow{1}{*}{acc} & \multirow{1}{*}{mAP} & \multirow{1}{*}{AUC} & \multirow{1}{*}{acc} & \multirow{1}{*}{mAP} & \multirow{1}{*}{AUC} & \multirow{1}{*}{acc} \\     
    \midrule
    AVFF & \textcolor{red}{\xmark} & $0.53$ & $0.54$ & $49.23$ & $0.54$ & $0.56$ & $56.78$ & $0.53$ & $0.52$ & $56.73$ & $0.53$ & $0.55$ & $53.85$    \\
     MRDF & \textcolor{red}{\xmark} & $0.58$ & $0.58$ & $54.81$ & $0.59$ & $0.60$ & $49.57$ & $0.57$ & $0.58$ & $55.71$ & $0.59$ & $0.59$ & $53.33$    \\    
    TALL & \textcolor{red}{\xmark} & $0.50$ & $0.50$ & $49.94$ & $0.47$ & $0.45$ & $43.60$ & $0.48$ & $0.48$ & $49.70$ & $0.47$ & $0.46$ & $46.38$    \\
      AVH-Align & \textcolor{red}{\xmark} & 0.56 & 0.57 & 55.18 & 0.51 & 0.51 & 42.28 & 0.58 & 0.60 & 57.82 & 0.54&  0.54 &  46.28  \\
    AVH-Align/sup & \textcolor{red}{\xmark} & 0.52 & 0.51 & 48.30  & 0.48 & 0.46 & 63.80 & 0.56 & 0.56 & 42.10 & 0.52 & 0.52 &   57.70 \\

    \midrule
       AVFF & \textcolor{ForestGreen}{\checkmark} & ${\color{blue}\mathbf{0.97}}$ & ${\color{blue}\mathbf{0.97}}$ & ${\color{blue}\mathbf{90.71}}$ & ${\color{blue}\mathbf{0.93}}$ & ${\color{blue}\mathbf{0.94}}$ & 
    ${\color{blue}\mathbf{85.10}}$ & ${\color{blue}\mathbf{0.93}}$ & ${\color{blue}\mathbf{0.93}}$ & ${\color{blue}\mathbf{86.58}}$ & ${\color{blue}\mathbf{0.91}}$ & ${\color{blue}\mathbf{0.93}}$ & ${\color{blue}\mathbf{84.85}}$    \\
    MRDF & \textcolor{ForestGreen}{\checkmark} & ${\color{RedOrange}0.90}$ & ${\color{RedOrange}0.92}$  & ${\color{RedOrange}79.55}$ & ${\color{RedOrange}0.75}$ & ${\color{RedOrange}0.81}$ & 
    ${\color{RedOrange}67.86}$ & ${\color{RedOrange}0.88}$ & ${\color{RedOrange}0.89}$ & ${\color{RedOrange}76.91}$ & ${\color{RedOrange}0.80}$ & ${\color{RedOrange}0.83}$ & ${\color{RedOrange}70.95}$ \\
    TALL & \textcolor{ForestGreen}{\checkmark} & $0.84$ & $0.85$ & $75.61$ & $0.67$ & $0.73$ & $62.40$ & $0.77$ & $0.78$ & $70.14$ & $0.69$ & $0.72$ & $63.86$    \\
      AVH-Align  & \textcolor{ForestGreen}{\checkmark} 
    & 0.56 &   0.56 & 52.48 & 0.51 & 0.51 & 62.66 & 0.57 & 0.57 & 58.11 & 0.54 & 0.54 &  57.78\\
    AVH-Align/sup  & \textcolor{ForestGreen}{\checkmark} 
    & 0.81 & 0.83 & 75.56 & 0.69 & 0.74 & 63.84 & 0.76 & 0.78  & 72.25 & 0.70 & 0.73 & 64.90 \\
    \bottomrule
  \end{tabular}
  }
    
    \label{table-results}
  \vspace{-0.2cm}
\end{table*}

We generate over 16,000 samples to cover both RVFA (real video, fake audio) and FVFA (fake video, fake audio) pairs.  We show the distribution of deepfakes across modalities  in Table~\ref{av-fake-distribution}. 
Manual inspection revealed that fake audio is generally easier to detect. Hence, we deliberately control the proportion of fake audio (FA) samples relative to fake videos (FV), ensuring that the dataset is sufficiently challenging for deepfake detectors. To create RVFA samples, we apply several VC models to some of the collected real videos. The samples produced by models from VC2 are reserved exclusively for our open-set model evaluation, and are not included in the training or validation subsets. The reference voices used for the VC models are sourced from Common Voice \cite{commonvoice:2020}, M-AILABS \cite{dataset2024m}, and VoxPopuli \cite{wang-etal-2021-voxpopuli}. To create FVFA samples, we apply two strategies. First, we generate talking-head videos using Sonic and Memo, using fake audio samples as input. These audio samples are created either via text-to-speech or voice conversion models. For the second method, we apply only voice conversion techniques to the audio tracks of already created synthetic videos. 

In Figure~\ref{fig_qualitative}, we present synthetic video frames produced by some of the deepfake methods. The samples are diverse and have a high degree of realism, confirming that MAVOS-DD represents a challenging dataset for existing deepfake detectors. For both real and generated videos, we highlight that the number of frames per second (FPS) ranges from $23$ to $60$. The audio bitrate varies between $88$ and $140$ \texttt{kbps}, with the audio sample rate spanning from $16$ to $44.1$ \texttt{kHz}. The video bitrate ranges from $40$ to over $10,000$ \texttt{kbps}.

\noindent
\textbf{Ethical considerations.}
We share MAVOS-DD under the International Attribution Non-Commercial Share-Alike 4.0 (CC BY-NC-SA 4.0) license, aiming for open and responsible research on deepfake detection. All real data samples are collected from public YouTube videos.  We would like to highlight that YouTube video datasets, such as YouTube-VOS \cite{Xu-ECCV-2018}, YouTube-VIS \cite{Yang-ICCV-2019}, and TalkingHead-1KH \cite{Wang-CVPR-2021b}, are widely popular in computer vision research. We verified that source channels do not explicitly forbid AI training. Since the videos are gathered from a public website, we adhere to the European Directive 2019/790\footnote{\url{https://eur-lex.europa.eu/eli/dir/2019/790/oj}} allowing researchers to use and store data from the public web domain for non-commercial research purposes. 
We thus store and provide access to the dataset for the purpose of verifying our research results. Access is restricted to non-commercial usage and will be granted upon completing a form. 
Moreover, we respect the individual privacy rights, including the right to be forgotten. If any individual identifies themselves in the dataset and wishes to have their data removed, they can contact us and we will promptly address the request by removing the respective video(s), in compliance with data protection principles.
\vspace{-0.1cm}
\section{Experiments}
\label{sec:experiments}
\vspace{-0.1cm}

\noindent
\textbf{Baselines and hyperparameters.} We conduct experiments using four state-of-the-art deepfake detectors. Three of them, namely AVFF~\cite{oorloff-CVPR-2024} MRDF~\cite{Zou-ICASSP-2024} and AVH-Align~\cite{Smeu-CVPR-2025} are multimodal, while the fourth one, TALL~\cite{xu-ICCV-2023}, analyzes only the video input. AVFF employs two unimodal encoders based on transformer blocks, each of them being trained to predict features of the opposite modality. The outputs from both encoders are concatenated and passed to a binary classifier for deepfake detection. Similarly, MRDF uses two encoders to extract features from each modality. The two encoders are based on ResNet-18~\cite{He-CVPR-2015}. Their output is concatenated and further processed by an audio-visual transformer module for deepfake detection. AVH-Align trains an audio-video feature alignment network, based on AV-HuBERT~\cite{Shi-ICLR-2022} features. TALL is a spatio-temporal modeling method that captures both spatial and temporal inconsistencies. The method is applicable to multiple architectures. In our work, we use TALL-Swin, which is based on Swin Transformer~\cite{liu-ICCV-2021}. We conduct the experiments using both pre-trained and fine-tuned versions of each model. We fine-tune MRDF for $5$ epochs, TALL for $15$ epochs, AVFF and AVH-Align for $10$ epochs on MAVOS-DD. The number of epochs are established based on early stopping. To optimize the models, we employ Adam~\cite{Kingma-ICLR-2015} with a learning rate of $10^{-3}$ for MRDF, $2\cdot10^{-5}$ for TALL, and $10^{-5}$ for AVFF and AVH-Align, respectively. We keep the default values for the other hyperparameters of Adam. We set the batch size to 4 for AVFF and MRDF, and 32 for TALL and AVH-Align. All the experiments are carried out on a computer with an Intel i9-14900K CPU with 192 GB of RAM and an Nvidia RTX 4090 GPU with 24 GB of VRAM.

\noindent
\textbf{Main results.} In Table~\ref{table-results}, we report the results for the four baseline models across three evaluation metrics: mean average precision (mAP), area under the ROC curve (AUC), and accuracy (acc). We report these values on all four test sets: in-domain, open-set model, open-set language and open-set full.

\begin{table}[t]
    \centering
    \caption{Results on BioDeepAV~\cite{Croitoru-Arxiv-2024} for pre-trained vs.~fine-tuned models.}
    \vspace{-0.15cm}
    \setlength\tabcolsep{0.28cm}
    \fontsize{8}{9}\selectfont{
    \begin{tabular}{lcccccc}
     \toprule
        \multirow{2.5}{*}{Model} & \multicolumn{3}{c}{Pre-trained} & \multicolumn{3}{c}{Fine-tuned on MAVOS-DD}  \\
        \cmidrule(lr){2-4}
     \cmidrule(lr){5-7}
        & mAP & AUC & acc & mAP & AUC & acc \\
    \midrule
    AVFF & 0.49 & 0.48 & 54.28 & 0.95 & 0.95 & 70.78 \\
    TALL & 0.55 & 0.49 & 44.69 & 0.83 & 0.82 & 74.66 \\
    MRDF & 0.52 & 0.52 & 48.56 & 0.53 & 0.53 & 52.12 \\
    \bottomrule

    \end{tabular}
    }
        \vspace{-0.1cm}

            \label{tab:biodeepav}
\end{table}

The results demonstrate that MAVOS-DD is a difficult data set for existing deepfake detection methods, since all the employed and publicly available pre-trained models perform close to random chance, regardless of the test set. We can attribute the performance gap of pre-trained models to the fact that MAVOS-DD typically contains examples that are more challenging to detect, since they are generated with models that exhibit a high degree of realism. The fine-tuned versions perform much better, especially in the in-domain scenario. With respect to the in-domain scenario, their performance levels decline in open-set setups, indicating that further developments are needed to improve the generalization of state-of-the-art detectors. As expected, the most significant performance drop is observed in the open-set model setup. This drop indicates that detectors still fail to generalize from a set of deepfake methods to another. The performance drop is lower in the open-set language case. However, when we examine the number of samples incorrectly predicted by the fine-tuned TALL model across in-domain and open-set language scenarios, we observe a difference of 5,917 samples, increasing from 2,999 to 8,916. A significant portion of these samples are likely mislabeled because the model only considers the video modality, being unable to detect RVFA samples by construction. Overall, there is a noticeable performance gap between the unimodal TALL method and the two multimodal approaches (AVFF and MRDF), suggesting that jointly analyzing visual and audio modalities provides a significant advantage on MAVOS-DD. 


\begin{figure}[t]
\centering
    \includegraphics[width=1.0\linewidth]{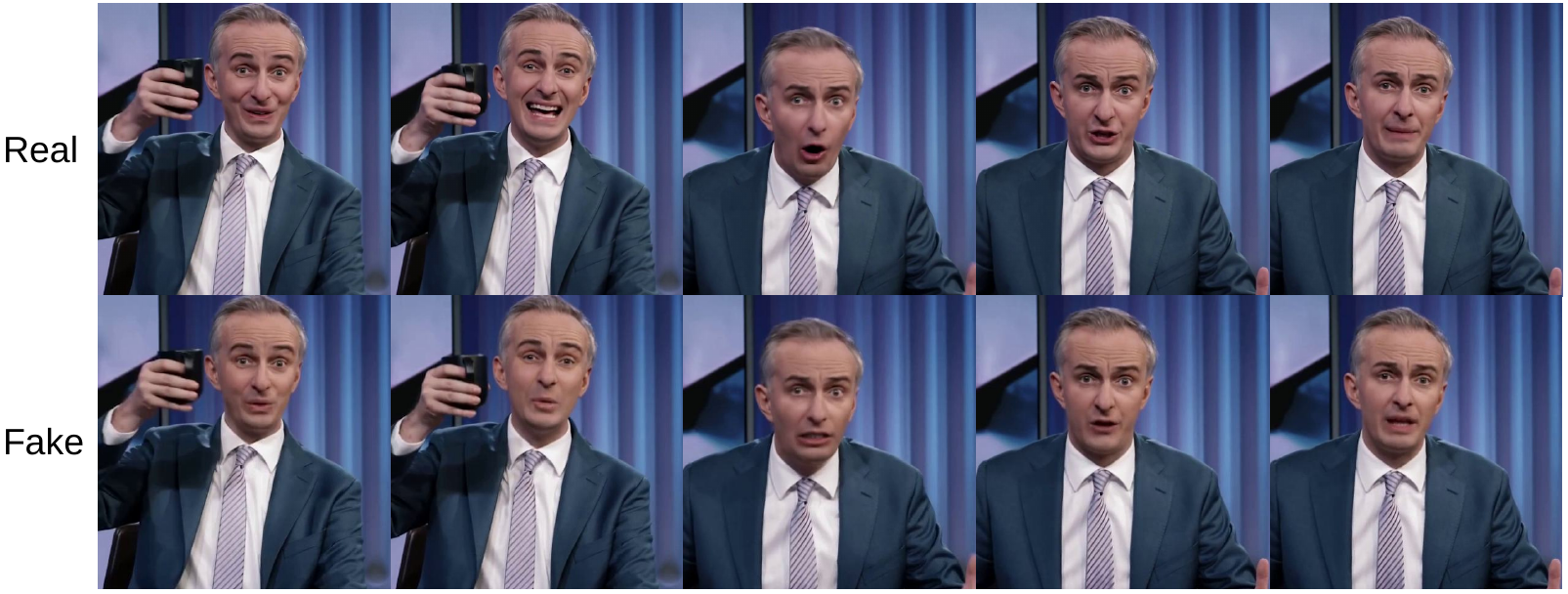}
    \vspace{-0.5cm}
    \caption{A real video and its corresponding fake sample generated using LivePortrait. The MRDF detector incorrectly classifies the fake sample as real. Best viewed in color.}
    \label{fig_liveportrait_sample}
    \vspace{-0.1cm}
\end{figure}

\begin{table}[t]
  \centering
  \caption{Cross-forgery experiments with AVFF and MRDF conducted on open-set full.}
    \vspace{-0.15cm}
  \setlength\tabcolsep{0.115cm}
  \fontsize{7.5}{8.5}\selectfont{
  \begin{tabular}{cccccccc}
    \toprule
    \multirow{2.5}{*}{Train forgery} & \multirow{2.5}{*}{Test forgery}  & \multicolumn{3}{c}{AVFF} & \multicolumn{3}{c}{MRDF} \\
    \cmidrule(lr){3-5}
    \cmidrule(lr){6-8}
     & & mAP & AUC & acc  & mAP & AUC & acc \\
    \midrule
    FVRA, RVFA, FVFA & FVFA & 0.99 & 0.99 & 93.31 & 0.92 &0.90 & 80.40\\
    FVRA, RVFA & FVFA & 0.98 & 0.98 & 92.73 & 0.88 & 0.88 & 78.04\\
    \bottomrule
  \end{tabular}
  }
  
  \label{table:cross-forgery}
\end{table}

\begin{table}
  \centering
      \caption{Results on MAVOS-DD per generative method. In the upper part, results correspond to in-domain methods. In the lower part, results correspond to cross-domain methods.}
  \setlength\tabcolsep{0.18cm}
  \vspace{-0.15cm}
  \fontsize{8}{9}\selectfont{
  \begin{tabular}{lcccccc}
    \toprule
    \multirow{2.5}{*}{Generative method} & \multicolumn{3}{c}{AVFF} & \multicolumn{3}{c}{MRDF}\\ 
     \cmidrule(lr){2-4}
     \cmidrule(lr){5-7}
     & mAP & AUC& Acc &  mAP & AUC& Acc \\
    \midrule
    Echomimic (EM) & 0.99 & 0.99 & 94.36 & 0.96 & 0.99 & 96.48\\
    Memo & 0.99 & 0.99 & 93.69 & 0.93 & 0.98 & 93.13\\
    LivePortrait (LP) & 0.98 & 0.98 & 90.85 & 0.46 & 0.41 & 50.16\\
    Inswapper & 0.84 & 0.83 & 74.96 & 0.89 & 0.90 & 83.17\\
    VC1 & 0.99 & 0.99 & 93.97 & 0.62 & 0.59 & 50.50 \\
    EM/Memo/LP + VC1 & 0.99 & 0.99 &  93.29 & 0.94 & 0.93 & 79.74\\
    \midrule
    Sonic & 0.99 & 0.99 & 93.55 & 0.94 & 0.97 & 90.08\\
    Hififace (HF) & 0.87 & 0.89 & 76.36 & 0.85 & 0.92 & 74.42\\ 
    Roop & 0.76 & 0.78 & 67.49 & 0.47 & 0.39 & 42.83 \\
    VC2 & 0.99 & 0.99 & 93.80 & 0.71 & 0.77 & 57.51\\
    Sonic/HF/Roop+VC2 & 0.99 & 0.99 & 93.31& 0.92 &0.90 & 78.04 \\
    \bottomrule
  \end{tabular}
  }
    \vspace{-0.25cm}

      \label{table-per-gen}
\end{table}

\begin{table*}[t]
  \centering
    \caption{Language-level results of AVFF on MAVOS-DD, under three scenarios. In the first scenario, videos contain the original audio. In the second scenario, audio is removed by replacing it with zero values. In the third scenario, audio is randomly replaced with that of another video.}
    \vspace{-0.15cm}
  \fontsize{8}{9}\selectfont{
  \begin{tabular}{lccccccccc}
    
    \toprule
      \multirow{2.5}{*}{Language} & \multicolumn{3}{c}{Correct audio} &\multicolumn{3}{c}{Missing audio}
  & \multicolumn{3}{c}{Mismatched audio}\\
  \cmidrule(lr){2-4}
     \cmidrule(lr){5-7}
     \cmidrule(lr){8-10}
  & mAP & AUC & acc & mAP & AUC & acc & mAP & AUC & acc \\
    \midrule
Arabic & 0.98 &	0.98 & 94.76 & 0.69 & 0.69 &	37.81 &	0.75 &	0.77 & 64.97 \\
English	&0.94&	0.94&	84.83&	0.61	&0.62	& 48.56  &	0.64 & 0.66& 61.53 \\
Mandarin & 0.93 & 0.96 & 90.48 & 0.70 & 0.74	& 67.26 & 0.69 & 0.74 & 69.94\\
Romanian &	0.99&	0.99	&97.87&	0.63&	0.64&	44.52	&0.81&	0.82&	70.90\\
Russian & 0.86 & 0.96 & 91.23 & 0.69 & 0.80	& 78.06 & 0.65 & 0.77 & 74.14\\
Spanish & 0.90 & 0.94 & 86.55 & 0.67 & 0.70	& 63.25 & 0.67 & 0.72 & 68.49\\
\midrule
German & 0.93 & 0.93 & 87.78 & 0.68 & 0.67	& 39.65 & 0.72 & 0.73 & 62.71 \\
Hindi & 0.79 & 0.78 & 76.13 & 0.59 & 0.57 & 	33.61 & 0.59 & 0.62 & 56.13 \\
    \bottomrule
  \end{tabular}
  }
  \label{table_language}
\vspace{-0.3cm}
\end{table*}

\noindent
\textbf{Cross-dataset experiment.} In Table~\ref{tab:biodeepav}, we evaluate the generalization benefits of fine-tuning on MAVOS-DD through a cross-dataset experiment. By testing both pre-trained and fine-tuned variants on BioDeepAV~\cite{Croitoru-Arxiv-2024}, we demonstrate that training on MAVOS-DD enhances the ability of these models to detect synthetic samples generated by other modern methods (NeRF, Gaussian Splatting) featured in BioDeepAV. 

\noindent
\textbf{Cross-forgery analysis.}
We conduct cross-forgery experiments by excluding the FVFA samples from the MAVOS-DD training set. We then evaluate models exclusively on this forgery type (FVFA), pairing it with an equally-sized set of real videos uniformly sampled from the \emph{open-set model} subset. The results are shown in Table~\ref{table:cross-forgery}, where we compare against models trained on the complete training set. The results indicate that the performance drops vary across models. AVFF \cite{oorloff-CVPR-2024} is more robust, showing minor performance drops, while MRDF \cite{Zou-ICASSP-2024} suffers greater declines in both mAP and AUC.

\begin{figure}[t]
    \centering
    \includegraphics[width=0.9\linewidth]{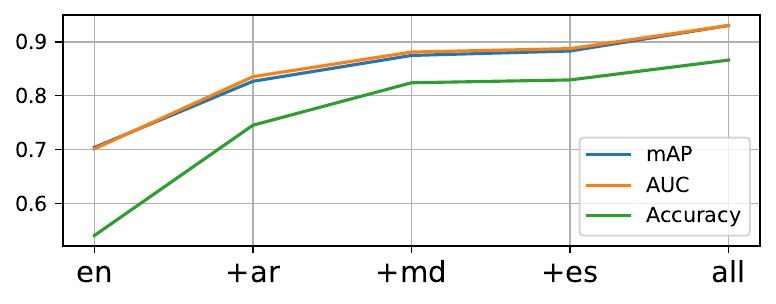}
    \vspace{-0.2cm}
    \caption{Open-set full performance of AVFF by progressively adding more languages to the training set. Best viewed in color.}
    \label{fig:progressive_train}
    \vspace{-0.35cm}
\end{figure}

\noindent
\textbf{Progressive training on languages.}
We explore the benefits of multilingual audio by progressively including more languages in the training data. We present the results in Figure~\ref{fig:progressive_train}. We can clearly observe that training on more languages boosts performance on the \emph{open-set full} split, as the extra languages improve the generalization capability of the AVFF model.

\noindent
\textbf{Results per generative method.} We report performance per generative method in Table~\ref{table-per-gen}, using balanced sets of real and fake videos for each case. Our analysis reveals that AVFF achieves near-perfect identification of synthetic audio samples in the open-set model scenario (last two rows). This is consistent with our initial observation that fake audio is sometimes easier to spot than fake video.


\noindent
\textbf{Error analysis.} We investigate which of the deepfake generative methods poses the greatest challenge for MRDF in terms of detection accuracy. We find that samples generated by LivePortrait and Roop are the most difficult, as detailed in Table~\ref{table-per-gen}. 
Roop is one of the methods included in the test set only, and we believe that this explains the poor performance of MRDF in identifying samples generated by Roop. In contrast, LivePortrait is part of the in-domain set, but the poor performance of the detector on this method can be attributed to the fact that we only synchronize the lips, leaving everything else as in the original video. In Figure~\ref{fig_liveportrait_sample}, we illustrate such a scenario where we show, side-by-side, frames from a real video and its corresponding fake video modified with LivePortrait. In the illustrated video, MRDF fails to detect the fake, misclassifying it as real.

\noindent
\textbf{Qualitative analysis.} 
To clarify the individual importance of audio and language features, we report language-level results. For a better assessment, we conduct two experiments. In the first setting, we remove the audio during inference by setting the audio features to zero. However, since this introduces a distribution shift between training and testing, we implement a second setting in which the audio for each video is randomly replaced with that of another video. This approach maintains a similar distribution of audio features and provides a more reliable assessment of the role played by the audio modality. We report the results for both experimental settings, along with the standard setup (where the correct audio is present) in Table~\ref{table_language}. All experiments are conducted using the open-set language setting.

The results reported in Table~\ref{table_language} emphasize the impact of the audio signal on the performance of AVFF. Moreover, proper synchronization between audio and video is essential, as demonstrated by the second experimental setting, where randomly replacing the audio results in a significant performance drop. If the model was not leveraging the interaction between audio and video, we would expect little to no impact from audio replacement. However, the observed decline in performance clearly indicates that the model does rely on the audio-video cues, underscoring the importance of the audio modality. 

\vspace{-0.1cm}
\section{Conclusion and Future Work}
\vspace{-0.1cm}

In this work, we introduced MAVOS-DD, a large-scale open-set benchmark for multilingual audio-video deepfake detection, comprising nearly 350 hours of real and generated videos. We further proposed a test split that creates four different evaluation scenarios: in-domain, open-set model, open-set language and open-set full. The resulting scenarios are aimed to assess the performance and robustness of deepfake detectors in challenging situations. We evaluated four different state-of-the-art deepfake detectors on the newly proposed benchmark, and observed significant performance drops across all four evaluation setups. The empirical results highlight the need to develop more robust deepfake detectors for practical scenarios.

In future work, we aim to continuously update the dataset by adding deepfake samples generated with models that are going to be released after our first release date. Thus, MAVOS-DD will keep up with the development pace of generative models, so that it will stay relevant for a long period of time. Additionally, we target the development of novel deepfake detectors that specifically address the challenges of the proposed open-set setups, which closely resemble real-world scenarios.  

\vspace{-0.1cm}
\section{Acknowledgments}
\vspace{-0.1cm}

This research is supported by the project ``Romanian Hub for Artificial Intelligence - HRIA'', Smart Growth, Digitization and Financial Instruments Program, 2021-2027, MySMIS no.~351416. This work is also supported by a grant of the Ministry of Research, Innovation and Digitization, CCCDI - UEFISCDI, project number PN-IV-P6-6.3-SOL-2024-2-0227, within PNCDI IV.

\bibliographystyle{IEEEtran}
\bibliography{IEEEabrv, refs}
\end{document}